\documentclass[final]{cvpr}

\usepackage{times}
\usepackage{epsfig}
\usepackage{graphicx}
\usepackage{amsmath}
\usepackage{amssymb}
\usepackage{subfiles}
\usepackage{multirow} 

\usepackage[pagebackref=true,breaklinks=true,colorlinks,bookmarks=false]{hyperref}

\def\cvprPaperID{1831} 
\def\confYear{CVPR 2021}

\begin{document}

\title{Few-Shot Domain Expansion for Face Anti-Spoofing}

\author{
Bowen Yang, Jing Zhang\\
College of Software, Beihang University\\
{\tt\small \{yang\_bw, zhang\_jing\}@buaa.edu.cn}
\and
Zhenfei Yin, Jing Shao\\
SenseTime Research\\
{\tt\small \{yinzhenfei, shaojing\}@sensetime.com}
}

\maketitle

\begin{abstract}
Face anti-spoofing (FAS) is an indispensable and widely used module in face recognition systems. Although high accuracy has been achieved, a FAS system will never be perfect due to the non-stationary applied environments and the potential emergence of new types of presentation attacks in real-world applications. In practice, given a handful of labeled samples from a new deployment scenario (target domain) and abundant labeled face images in the existing source domain, the FAS system is expected to perform well in the new scenario without sacrificing the performance on the original domain. To this end, we identify and address a more practical problem: \textbf{Few-Shot Domain Expansion} for Face Anti-Spoofing (FSDE-FAS). This problem is challenging since with insufficient target domain training samples, the model may suffer from both overfitting to the target domain and catastrophic forgetting of the source domain. To address the problem, this paper proposes a \textbf{Style transfer-based Augmentation for Semantic Alignment} (SASA) framework. We propose to augment the target data by generating auxiliary samples based on photorealistic style transfer. With the assistant of the augmented data, we further propose a carefully designed mechanism to align different domains from both instance-level and distribution-level, and then stabilize the performance on the source domain with a less-forgetting constraint. Two benchmarks are proposed to simulate the FSDE-FAS scenarios, and the experimental results show that the proposed SASA method outperforms state-of-the-art methods.

\end{abstract}

\section{Introduction}

\begin{figure}[htb]
\centering
\includegraphics[width=1\linewidth]{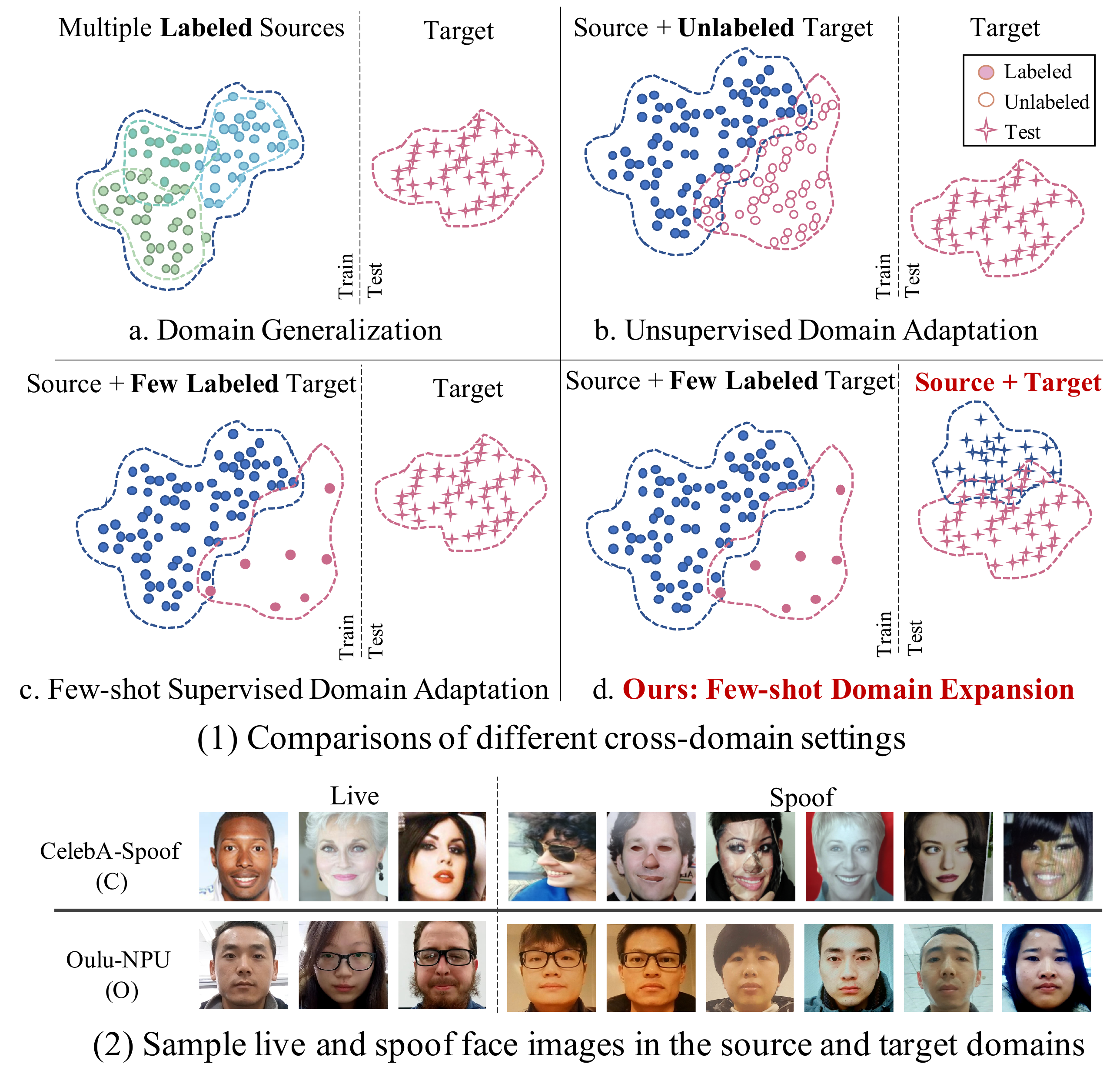}
\caption{
    The proposed Few-Shot Domain Expansion for Face Anti-Spoofing (FSDE-FAS) aims at expanding the source domain with only a few target domain samples, hoping for improving the performance of FAS in the target domain without sacrificing the source domain performance.
}
\label{figure:illustration}
\vspace{-2em}
\end{figure}


Nowadays, face recognition has been widely used in many interactive AI systems, such as cell phone unlocking, access control, and security check. 
Wherein, Face anti-spoofing (FAS) algorithms play an important role to ensure the security by preventing face abuse.
Recent years, various advanced face anti-spoofing methods~\cite{boulkenafet2016face, wen2015face,erdogmus2014spoofing,pinto2015face}, especially the deep neural networks (DNNs)-based ones~\cite{cai2020drl,liu2020disentangling,wang2020deep,yu2020searching}, are proposed with promising performance on the intra-domain setting.
However, in a real world FAS system such as mobile face unlock, the applied environments are non-stationary.
Different presentation attack (PA) types or scenarios out of training distribution~\cite{arashloo2017anomaly} may appear, leading to performance degradation.
We have to collect and annotate a large amount of training data for new emerging domains, which is labour intensive and expensive.
By contrast, a more common fact is that only few labeled target domain samples are available before deployment~\cite{qin2019learning}. 
To meet the practical application requirements, the FAS model should be able to adapt the target environments and PA types with few examples, without forgetting the already learned counterparts in the source domain. Thus, the problem is not simply a domain adaptation~\cite{ganin2015unsupervised} (DA), domain generalization~\cite{shao2019multi,jia2020single} (DG), or few-shot learning~\cite{qin2019learning} (FSL) problem, since DA, DG, and FSL all merely focus on improving the performance of the new target domains or tasks (shown in the upper half of Figure~\ref{figure:illustration}), while the practical requirement of face anti-spoofing is to EXPAND the source domain to a more diverse compound domain to handle different environments and PA types (shown in the low half of Figure~\ref{figure:illustration}). 

To this end, this paper defines and tackles the problem of Few-Shot Domain Expansion for Face Anti-Spoofing (FSDE-FAS). That is, with limited target domain data as well as abundant source data, the goal of FSDE-FAS is to improve the performance of FAS in the target domain without performance degradation in the source domain. In other words, the model should perform well on the expanded joint domain composed of both the source domain and the target domain. 
The task is important yet challenging. As we will show later through experiments, the target domain performance can hardly be improved sufficiently by naive joint training with insufficient target domain labeled samples. Conversely, the cross-domain methods (e.g. DA, DA, and FSL) may bring improvements in the target domain but lead to detrimental effects in the source domain. 

\begin{figure}[t]
\centering
\includegraphics[width=0.47\textwidth]{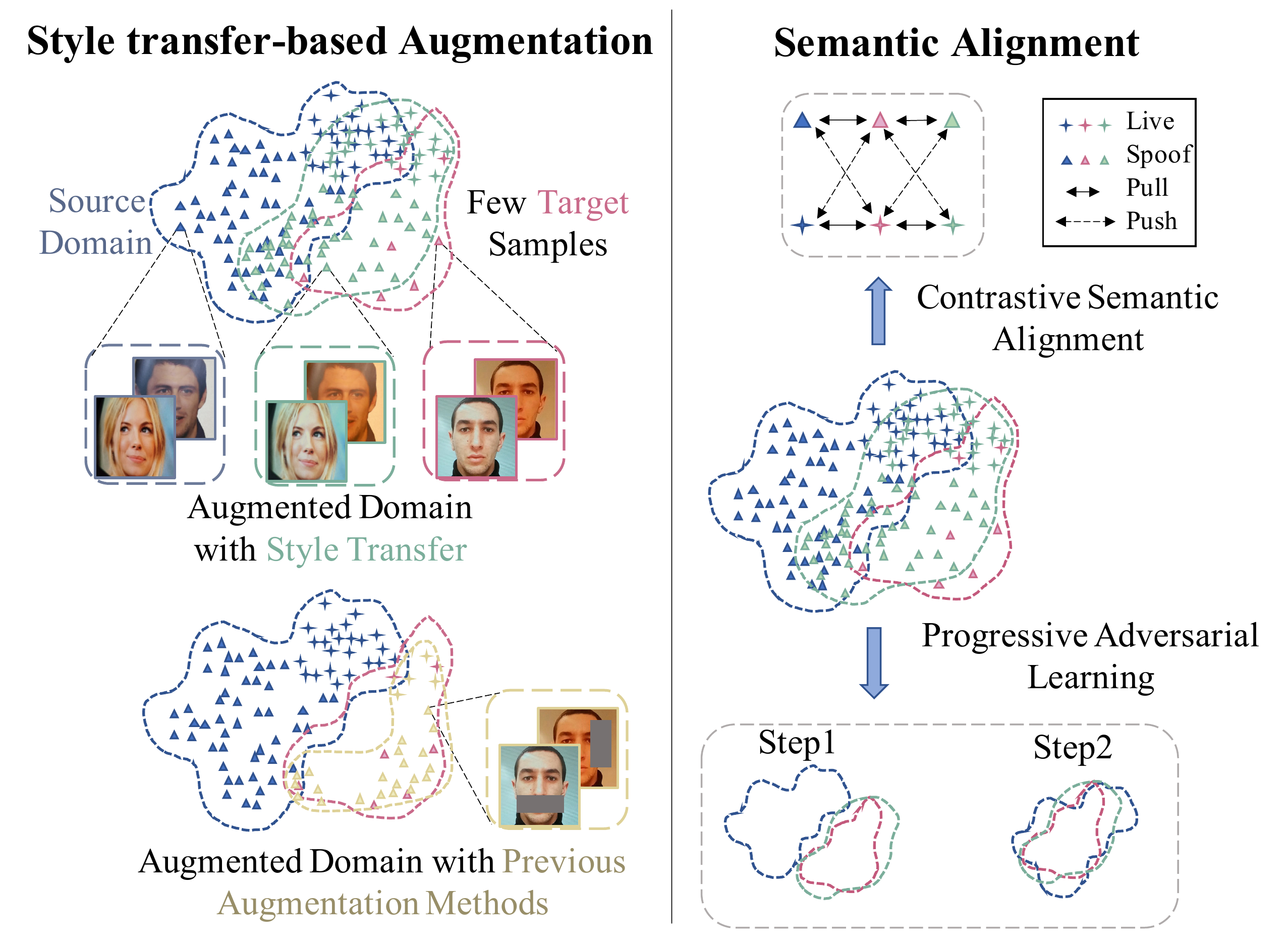}
\caption{
     Left: 
    Illustration of the comparisons between the proposed style transfer-based augmentation method and other general augmentation methods. The style transfer-based method aims at generating samples lie between the source and target domains while preserving the semantic information.
    Right: With the help of auxiliary data, the domain divergences are reduced from both instance-label via contrastive semantic alignment and distribution-level by a progressive adversarial training strategy.
}
\label{figure:wct2}
\vspace{-1em}
\end{figure}

To tackle such a problem, we propose a Style Augmented Semantic Alignment (SASA) framework, which create an auxiliary domain by \textit{augmenting} the target domain data and then smoothly \textit{align} the target data to the source domain with the assistant of the auxiliary domain. 
As shown in Figure~\ref{figure:wct2}, we argue that the commonly used data augmentation methods~\cite{krizhevsky2017imagenet,wan2013regularization,zhong2020random} for general image classification are not directly applicable, because in our setting, the desired augmented data should be able to characterise the target domain discriminative information, preserve the source domain performance, and bridge the source and target domains, simultaneously. In consideration of these properties, we propose a novel augmentation method for our FSDE-FAS problem based on style transfer. The motivation is that the distribution divergence between the source and target domains in FAS are mainly from two levels: 1) low-level visual shift caused by the changing environments, device parameters, or different PA types, and 2) high-level semantic shift due to different identities and global arrangement of images. 
Therefore, naturally, we propose to construct the auxiliary domain by generating synthetic samples with source image contents and target image styles based on a photorealistic style transfer method WCT2~\cite{yoo2019photorealistic}, because the style features contain more low-level texture and color information and the content features contain the high-level identity and global arrangement in a face image. 
In such a way, the generated images constitute a desired auxiliary domain that well meet our requirements.



After obtaining the augmented auxiliary domain, we are ready to align different domains to further improve the new domain performance. Here, we propose to use contrastive semantic alignment and progressive adversarial learning to align different domains from both instance-level and distribution-level. Specifically, the contrastive semantic alignment aims at learning more compact representations within each class and encouraging class separation between different domains at the instance-level. Note that we do not apply the contrastive semantic alignment to the samples within the intra-domain to avoid destroying the potential semantic distribution and hurt the source domain performance. By considering that merely aligning different domains from the instance-level may be prone to overfitting to specific instances, we further align the whole distributions using adversarial learning, thanks to the additional augmented auxiliary domain data that characterise the target domain semantic distribution to a certain degree. Concretely, we first align the distributions of the target and auxiliary domain, and then treating them as a whole and further align its distribution with the source domain.
To maximally preserve the source domain performance, we penalize the shift of source data in the feature space during training. In this way, the target domain is smoothly aligned to the source domain with the assistant of the augmented domain.
The main contributions of this paper are:
\begin{itemize}
\item [1)] 
Based on the real world requirements in FAS, we define and address a practical and challenging problem, named few-shot domain expansion in FAS (FSDE-FAS) for the first time. With a handful of target domain samples and plenty of source domain data, the goal of FSDE-FAS is to learn a robust model for the expanded joint domain of the source domain and target domain.
\item [2)]
To tackle the FSDE-FAS problem, we propose a Style transfer-based Augmentation for Semantic Alignment (SASA) framework to augment the target domain with a synthesized auxiliary domain through style transfer, and then propose a carefully designed domain alignment mechanism to improve the performance on the expanded domain.
\item [3)]
Two benchmarks are created for the proposed FSDE-FAS setting, and extensive experiments are conducted to verify the effectiveness of the proposed method.
\end{itemize}

\section{Related Work}

\subsection{Face Anti-spoofing Methods}
The technical exploration of face anti-spoofing can be roughly divided into two categories: appearance-based methods and temporal-based methods. Appearance-based methods are mainly dedicated to mining the subtle texture difference between live and spoof caused by the difference between human skin and spoof materials. Researchers mainly extract handcrafted features from the faces, e.g., LBP~\cite{boulkenafet2015face,de2012lbp,de2013can}, HoG~\cite{komulainen2013context,yang2013face}, SIFT~\cite{patel2016secure} and SURF~\cite{boulkenafet2016face}, and train a classifier to discern the live vs. spoof, such as SVM and LDA. 
In recent years, deep learning methods have shown remarkable advantages. Methods in \cite{li2016original,patel2016cross,feng2016integration} split live and spoof by training a deep neural network with binary classifier. Moreover, some useful auxiliary supervision are introduced to improve feature learning, such as depth map~\cite{atoum2017face} and reflection map~\cite{kim2019basn}.
Differently, temporal-based methods aim to discover spoof trace via extracting temporal cues through multiple frames. \cite{liu20163d,liu20163d} estimate rPPG signals from RGB face videos as attack clue. \cite{Siw} proposed a CNN-RNN model to use face depth and rPPG signals, simultaneously. However, most of these methods fail to explicitly consider the generalization capacities to a new domain when a handful of new domain samples are available. It is also noteworthy that our method mainly focus on improving the performance on the expanded domain through data augmentation and domain alignment, and thus is orthogonal to these advanced general FAS methods and other training strategies (e.g. meta-learning).

\subsection{Cross-domain Methods}
Many cross-domain settings have been proposed to deal with the issue of distribution divergence between different domains~\cite{saenko2010adapting,shimodaira2000improving}, such as domain adaptation~\cite{ganin2015unsupervised,motiian2017unified,motiian2017few} (DA), domain generalization~\cite{blanchard2011generalizing,muandet2013domain,khosla2012undoing,xu2014exploiting} (DG) and domain expansion~\cite{jung2017less} (DE).
Our assumptions about data availability are the same as \cite{motiian2017unified,motiian2017few}, both of which proposed to effectively improve the performance on target domain with abundant source domain labeled data and only a few target domain labeled data. However, they generally ignore the negative impact on the source domain performance after adaptation.
Our expectations for the model performance are similar to \cite{jung2017less}, which formally defined the concept of domain expansion. The difference is that \cite{jung2017less} assumes that there is no access to the source domain data, but the target domain has a large amount of labeled data. By contrast, we do not limit the access to the source data, but only assume to have a few target data, which is consistent with the real world FAS scenarios.
Domain generalization assumes that there exists a generalized feature space underlying the seen multiple source domains and the features in this space can generalize well on the unseen but related target domains.
Recently, several works~\cite{shao2019multi,jia2020single} introduce DG approaches into FAS to learn a general feature expression. However their methods can not be used effectively when the available multiple source domains are imbalanced in terms of sample numbers (i.e. in the few-shot cases).
AIM-FAS~\cite{qin2019learning} formulates FAS as a zero- and few-shot learning problem and proposes an adaptive inner-update meta-learning strategy which also only focus on target domain performance.
Our experimental results show that the previous cross-domain methods may potentially sacrifice the performance on the source domain, which is undesirable in FAS task. This motives our introduction of domain expansion for FAS.

\section{SASA Framework}
\begin{figure*}[htb]
\centering
\includegraphics[width=0.92\textwidth]{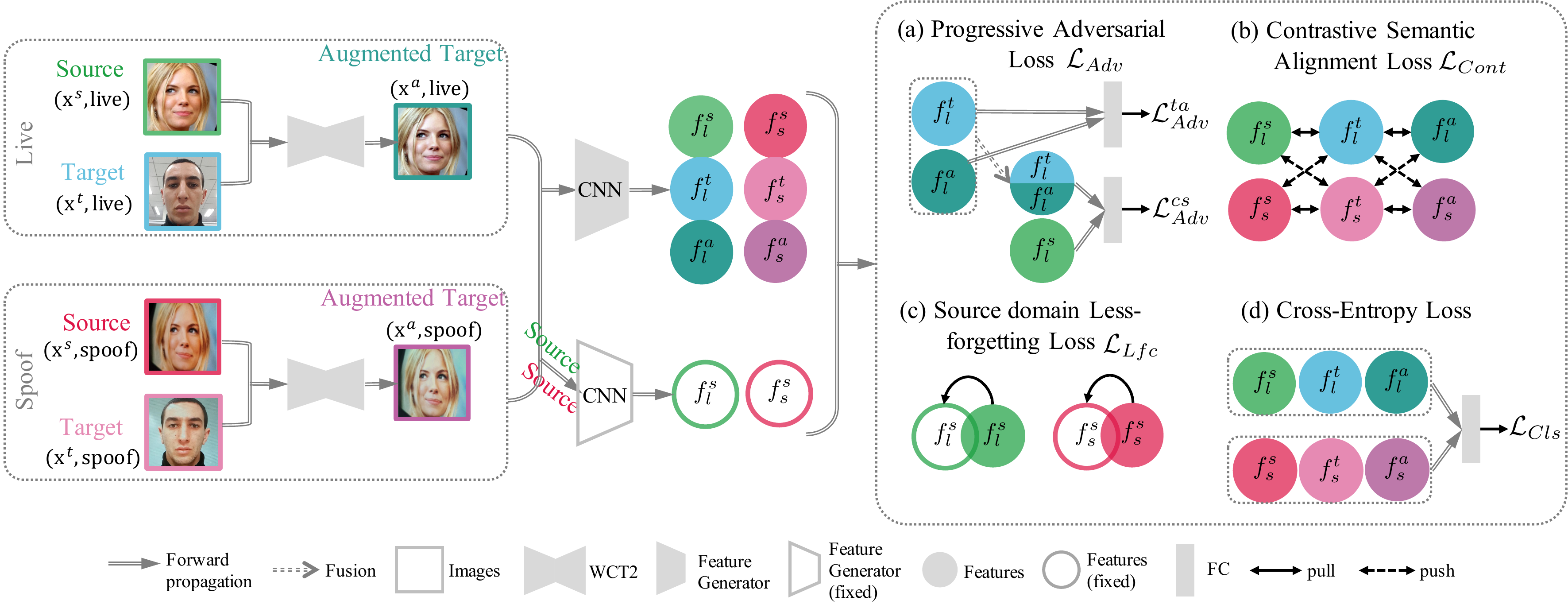}
\caption{An overview of the proposed Style transfer-based Augmentation for Semantic Alignment (SASA) framework. To deal with the issue of insufficient target training samples, the photorealistic image stylization method WCT2~\cite{yoo2019photorealistic} is used to generate auxiliary domain samples with source contents and target styles. 
With the generated samples, different domains are then aligned from both instance-level through contrastive semantic alignment and distribution level via a progressive adversarial training strategy.
The source domain less-forgetting loss is applied to source features to stabilize the performance on the source domain. Superscript indicates the domain to which the sample belongs: source (s), target (t), augmented (a). Subscript of features $f$ indicates the class labels (i.e. l: live, and s: spoof).
}
\label{figure:network}
\vspace{-1em}
\end{figure*}

In this section, we present our method in details.
As illustrated in Figure~\ref{figure:network}, we propose a Style Augment Semantic Alignment (SASA) framework for training the face anti-spoofing model, such that it performs well on the joint expanded domain with both the source and the target data. Firstly, a photorealistic image stylization method is used on pairs of images from source and target domains with the same class to synthesize new images, constituting an augmented auxiliary domain. Then, the adaptability on the target domain is further improved with the help of the auxiliary domain. We propose to reduce the distribution divergences between different domains from both instance-label via a contrastive semantic alignment method and distribution-level through a progressive adversarial training strategy. We argue that the two levels of distribution alignment are complementary to each other.
Additionally, we stabilize the performance on source domain by penalizing the shift of source data in the feature space.

To begin with, we give the definitions of the notations. We denote an existing dataset with abundant training samples as $\mathcal{D}_{s}= \left \{ \left ( \mathbf{x}^{s}_{i},y^{s}_{i} \right ) \right \}_{i=1}^{N}$ and a new coming dataset with only a few samples as $\mathcal{D}_{t}= \left \{ \left ( \mathbf{x}^{t}_{j},y^{t}_{j} \right ) \right \}_{j=1}^{M}$, where $\mathbf{x}^{s}_{i}$ ($\mathbf{x}^{t}_{j}$) is the $i$-th ($j$-th) labeled source (target) sample with label $y^{s}_{i}$ ($y^{t}_{j}$), and $M\ll N$. 
We assume that there is a covariate shift~\cite{shimodaira2000improving} between $X^{s}=\{\mathbf{x}^s_i\}$ and $X^{t}=\{\mathbf{x}^t_j\}$ due to the changing environments and PA types, i.e. there is a divergence between the marginal distributions $p(X^{s})$ and $p(X^{t})$. 
Different from traditional domain adaptation, our goal is to learn a prediction function $F = H \circ G:\mathcal{X}\rightarrow \mathcal{Y}$ 
that is expected to perform well on data from both the source domain and the target domain, simultaneously. Here, $G$ refers to the feature generator, $H$ refers to the live/spoof classifier, and $\mathcal{Y}=\{0,1\}$ in FAS problems.
With abundant source domain data, it is reasonable to assume that a well-trained source model feature extractor $G_{0}$ with sufficiently good classification capacities on the source domain is available.

\subsection{Style Transfer-based Data Augmentation}

To deal with the issue of data insufficiency in the target domain, we propose a data augmentation method based on style transfer. 
We utilize the state-of-the-art photorealistic image stylization method WCT2~\cite{yoo2019photorealistic} to generate auxiliary images using source domain images as content input and target domain images as style input.
The procedure of generating data can be formulated as:
\begin{equation}
\label{equation:STDA}
    \mathbf{x}^{a}_{k} = \mathcal{F_{ST}}\left (  \mathbf{x}^{s}_{i},\mathbf{x}^{t}_{j}\right )\ ,
\end{equation}
where $\mathcal{F_{ST}}$ is the stylization transform net proposed in~\cite{yoo2019photorealistic} with the $\mathbf{x}^{s}_{i}$ as content input and the $\mathbf{x}^{t}_{j}$ as style input, and $\mathbf{x}^{s}_{i}$ and $\mathbf{x}^{t}_{j}$ are samples of the same category randomly selected from the source domain and the target domain, respectively, i.e. $y^{s}_{i} = y^{t}_{j}$. The generated data $\{ \mathbf{x}^{a}_{k} \}$ forms a new auxiliary domain $X^{a}$, which lies between $X^{s}$ and $X^{t}$.

In the auxiliary domain, the live (hack) images are composed of the content of the source domain live (hack) images and the style of the target domain live (hack) images. Hence, the images in the auxiliary domain have the same identity information as in the source domain and similar low-level features (e.g. textures and colors) with the target domain. For the augmented live images, the low-level features encode more information about the target domain lighting conditions or camera parameters, while for the augmented hack images, the spoof traces are also encoded. Therefore, to further improve the performance on the joint domain, we still need to carefully align different domains. In the following two sections, we present two methods to align different domains from both instance-level and distribution-level.

\subsection{Contrastive Semantic Alignment}
The underlying distribution of the target domain can hardly be characterised by only a few samples, making it sub-optimal for directly aligning the target sample distribution with the source one. Alternatively, measuring the pairwise distances between samples in the feature space is a more feasible option. 
We propose to use contrastive semantic alignment method by manipulating pairwise distances to achieve semantic alignment for positive pairs and separation for negative pairs as follows,

\begin{footnotesize}
\begin{equation}
\label{equation:loss_semantic_alignment}
    \mathcal{L}_{Sem}(G) = \sum_{i,j}\frac{1}{2}\left \| G(\mathbf{x}^{s}_{i})-G(\mathbf{x}^{t}_{j}) \right \|^{2} + 
    \sum_{j,k}\frac{1}{2}\left \| G(\mathbf{x}^{t}_{j})-G(\mathbf{x}^{a}_{k}) \right \|^{2}\ ,
\end{equation}

\begin{equation}
\label{equation:loss_separation}
\begin{split}
    \mathcal{L}_{Sep}(G) = &\sum_{i,j}\frac{1}{2}\mathrm{max}(0, m-\left \| G(\mathbf{x}^{s}_{i})-G(\mathbf{x}^{t}_{j}) \right \|^{2}) +\\ 
    &\sum_{j,k}\frac{1}{2}\mathrm{max}(0, m-\left \| G(\mathbf{x}^{t}_{j})-G(\mathbf{x}^{a}_{k}) \right \|^{2})\ ,
\end{split}
\end{equation}
\end{footnotesize}
where in Eq.~\ref{equation:loss_semantic_alignment}, $y_{i}^{s} = y_{j}^{t} = y_{k}^{a}$, while in Eq.~\ref{equation:loss_separation}, it is assumed that $y_{i}^{s} \neq  y_{j}^{t}$ and $y_{j}^{t} \neq y_{k}^{a}$. $\|\cdot\|$ denotes the Frobenius norm, and $m$ is the margin that specifies the separability in the embedding space.

The advantage of this approach is that it makes full use of the target data by allowing even a single target sample to be paired with all the source samples. 
Note that we only pair the auxiliary domain with the target domain but not the source domain, since the images in the auxiliary domain contain more target domain specific spoof traces carried in the target styles. If the contrastive loss is applied between the auxiliary domain and the source domain, the discriminative information learned from the target spoof traces will be compromised. 
Moreover, we only sample pairs between different domains, not within each individual domain, such that the data structure in each domain and the representation capabilities of $g$ are both maximally preserved. Because spoof is an abstract collection of sub-concepts, including different spoof types such as print and replay, as well as various fine-grained spoof materials. 
Deliberately driving the features of intra-class closer within domain may destroy the underlying semantic distribution.

Finally, we obtain total contrastive semantic alignment loss by adding two parts together:
\begin{equation}
\label{equation:loss_contrastive}
    \mathcal{L}_{Cont} = \mathcal{L}_{Sem}(G) + \mathcal{L}_{Sep}(G)\ .
\end{equation}

\subsection{Progressive Adversarial Learning}

Contrastive semantic alignment works at the instance-level, which can effectively narrow the distance between different domains, but cannot fully align the distributions. Benefiting from the additional augmented domain data, we design a two-step strategy to align the distribution of $X^{t}$ with $X^{a}$, and then align the distribution of their combination $X^{c}$ with $X^{s}$ using progressive adversarial learning,

\begin{footnotesize}
\begin{equation}
\label{equation:loss_adv_ta}
\begin{split}
    \min_{D_{ta}}\max_{G}\mathcal{L}_{Adv}^{ta}(G,D_{ta})=&\mathbb{E}_{\mathbf{x}\sim \{X^{t},X^{a}\}}[\log{D_{ta}(G(\mathbf{x}))} ] + \\&\mathbb{E}_{\mathbf{x}\sim \{X^{t},X^{a}\}}[\log({1-D_{ta}(G(\mathbf{x}))}) ]\ ,
\end{split}
\end{equation}

\begin{equation}
\label{equation:loss_adv_tas}
\begin{split}
    \min_{D_{cs}}\max_{G}\mathcal{L}_{Adv}^{cs}(G,D_{cs})=&\mathbb{E}_{\mathbf{x}\sim \{X^{c},X^{s}\}}[\log{D_{cs}(G(\mathbf{x}))} ] + \\&\mathbb{E}_{\mathbf{x}\sim \{X^{c},X^{s}\}}[\log({1-D_{cs}(G(\mathbf{x}))}) ]\ ,
\end{split}
\end{equation}
\end{footnotesize}
where different domain discriminators are used to compete against the feature generator $G$ between pairs of domains. Specifically, in the early stage of training, only the discriminator $D_{ta}$ is trained to learn a shared and discriminative feature between $X^{t}$ and $X^{a}$ by optimizing $\mathcal{L}_{Adv}^{ta}$. The justification is that, the distributions of $X^{t}$ and $X^{a}$ lie closer in the feature space since they are designed to have the similar styles which encode more discriminative information. By aligning them first will not destroy much structural and semantic information in the target domain. After $D_{ta}$ is sufficiently trained, features of face images from $X^{t}$ and $X^{a}$ are indistinguishable. Then we consider $X^{t}$ and $X^{a}$ as a whole $X^{c}$ and train the discriminator $D_{cs}$ to encourage the feature from $X^{c}$ and $X^{s}$ to be indistinguishable by optimizing $\mathcal{L}_{live}^{tas}$. 
The progressive adversarial loss is formed by combining the two domain losses,
\vspace{-0.3em}
\begin{equation}
\label{equation:loss_adv}
    \mathcal{L}_{Adv} =\mathcal{L}_{Adv}^{ta}(G,D_{ta}) + \mathcal{L}_{Adv}^{cs}(G,D_{cs})\ .
\end{equation}

We argue that the progressive strategy can align the target domain with the source domain more smoothly. 
During the early stage of training, the target domain moves towards the intermediate auxiliary domain first. And then the second stage can make better use of the augmented data, which have domain-invariant features as the target domain, to better mimic the target domain distribution.
\subsection{Source Domain Less-forgetting Constraint}
Since the augmented data are composed of features of the source domain and the target domain, the discriminative spoof traces in the source domain are also carried in the augmented data. Thus, the introduction of the augmented auxiliary domain can already preserve the source domain performance to a certain degree. However, it may not be sufficient to fully alleviate the less-forgetting effects without an explicit constraint. Thus, to further maintain the stability of the performance on the source domain, a less-forgetting constraint is further incorporated in the training process. 

We treat the pre-trained source model as a fixed teacher model, and the adapted model as a student model. During training the student model, the predictions of the source domain samples should not be changed much. Here, we use a mean square error loss to penalize the excessive shift of the source features, which is formulated as follows,
\vspace{-0.1em}
\begin{equation}
\label{equation:loss_contrastive}
    \mathcal{L}_{Lfc}(G) =  \sum_{x\in X^{s}} \| G(\mathbf{x})-G_{0}(\mathbf{x})  \| _{2}^{2}\ .
\end{equation}

\subsection{Overall Objective Function}
The final objective is a combination of the above-mentioned losses, together with the cross-entropy loss $\mathcal{L}_{Cls}$ on all the labeled samples from the source, target, and the auxiliary domain. 
\begin{equation}
\label{equation:lentropyoss_contrastive}
    \mathcal{L}_{total} =  \mathcal{L}_{Cls} + \lambda _{1}\mathcal{L}_{Cont} + \lambda _{2}\mathcal{L}_{Adv} + \lambda _{3}\mathcal{L}_{Lfc}\ ,
\end{equation}
where $\lambda_{1}$, $\lambda_{2}$ and $\lambda_{3}$ are trade-off parameters.

\begin{table}[b]
\vspace{-0.3em}
\begin{center}
\caption{Specifications of four FAS datasets used in this paper.}
\vspace{3pt}
\resizebox{0.47\textwidth}{!}{
\label{table:datasets}
\begin{tabular}{cccccc}
\hline
Dataset  & Subjects & Data(V/I) & Sensor & Spof Type         & Session \\ \hline
CelebA-Spoof & 10177 & 625,537 (I) & \textgreater{}10 & \begin{tabular}[c]{@{}c@{}}3 Print, 3 Replay,\\ 1 3D, 3 Paper Cut\end{tabular} & 8 \\
MSU-MFSD & 35       & 440 (V)   & 2      & 1 Print, 2 Replay & 1       \\
Oulu-NPU & 55       & 5,940 (V) & 6      & 2 Print, 2 Replay & 3       \\
SiW      & 165      & 4,620 (V) & 2      & 2 Print, 4 Replay & 4       \\ \hline
\end{tabular} }
\end{center}
\vspace{-1em}
\end{table}

\section{FSDE-FAS Benchmarks}
We select four publicly available face anti-spoofing datasets to evaluate the effectiveness of our method:
Celeba-Spoof~\cite{CelebA-Spoof} (C for short), MSU-MFSD~\cite{MSU-MFSD} (M for short), Oulu-NPU~\cite{Oulu-NPU}  (O for short), SiW~\cite{Siw} (S for short).
Table~\ref{table:datasets} shows the specifications of the four datasets.
Noticing that Celeba-Spoof has far more subjects and variations than any other datasets, a model trained on which can better simulate a real face anti-spoofing system. Therefore, we choose Celeba-Spoof as the source domain and treat the other three datasets as three target domains.
We design two benchmarks to evaluate the adaptability (improving performance on new domains) and stability (preserving performance on origin domains) of our method for FSDE-FAS.

{\bf CMOS-ST} is a single target domain benchmark built based on Celeba-Spoof and three target domain datasets, respectively, (i.e. C $\rightarrow$ M, C $\rightarrow$ O and C $\rightarrow$ S).
For each of the three tasks, we generate our training set with the whole training set of Celeba-Spoof and samples of \textbf{one} subject random selected from the target domain training set.
In particular, one random frame of all videos including the selected subject are contained (12 frames for MSU-MFSD, 90 frames for Oulu-NPU, and 28 frames for SiW ).
Such a design ensures complete coverage using least target training samples.
The whole test set of Celeba-Spoof and the rest of the samples in the target domain will be used for evaluation.

{\bf CMOS-MT} is a multi-target domain benchmark built based on Celeba-Spoof and all of the three target domain datasets, (i.e. C $\rightarrow$ M\&O\&S). 
The training and test sets are defined similarly as in {\bf CMOS-ST}.
The only difference is that the training and test data of the three target domains will be used simultaneously.
This benchmark is designed to evaluate whether the proposed method can handle interactions between multiple target domains.

By following the cross-dataset evaluation protocols for FAS proposed in~\cite{CelebA-Spoof}, We employ Attack Presentation Classification Error Rate (APCER), Bona Fide Presentation Classification Error Rate (BPCER), and ACER~\cite{standard2016information} to evaluate the performance stability on the source domain (Celeba-Spoof) and Half Total Error Rate (HTER)~\cite{de2013can} to validate the adaptability on target domains.
Note that, in the evaluation of the target domains, we assume that no validation set is available and use the same threshold based on source validation set.
In order to ensure the stability and reliability of the results, we report the final results of each set of experiments by averaging five different runs.

\section{Experiments}


\subsection{Implementation Details}
For data pre-processing, we use MTCNN~\cite{zhang2016joint} to detect and align face and resize all the detected faces to $256\times 256\times 3$ for source and target samples. 
Then we randomly sample source domain samples that provide the contents and the target samples that provide the styles to synthesize the augmented data using WCT2~\cite{yoo2019photorealistic}. The number of the synthesized samples is about one tenth of the source.
We use ResNet-18~\cite{he2016deep} as the backbone architecture of feature generator and a two nodes FC layer as the classifier. Domain discriminator contains two FC layers with 512 and 2 nodes, respectively. The Adam optimizer with weight decay of 1e-5 and learning rate of 1e-4 is used for optimization. The hyperparameters $\lambda_{1}$, $\lambda_{2}$ and $\lambda_{3}$ are set to 1e-3, 1.0 and 10, respectively. During training, we force every batch to read the same number of live and spoof images, with 64 from $X^{s}$, 4 from $X^{t}$ and 8 from $X^{a}$. We implement $\mathcal{L}_{Adv}$ on live images only based on the observation that the spoof images bring little performance improvements by applying $\mathcal{L}_{Adv}$, which is consistent with the observations in \cite{jia2020single}. Please see Supplementary for more details.

\subsection{Experiments on FSDE-FAS}

To validate the performance of SASA on few-shot domain expansion problem, we compare SASA with representative general cross-domain recognition methods and FAS mathods which are closely related to our setting or approach.
\textbf{CCSA}~\cite{motiian2017unified} and  \textbf{FADA}~\cite{motiian2017few} focus on solving general supervised domain adaptation (SDA) problems, especially when only few target training samples are available; \textbf{SSDG}~\cite{jia2020single} deals with the domain generalization face anti-spoofing (DG-FAS) problem and achieves state-of-the-art performance. All of these methods can fulfill partial but not all the goals in FSDE-FAS. It is worth mentioning that although part of our task is similar to AIM-FAS~\cite{qin2019learning}, 
which focus on zero-shot and few-shot face anti-spoofing. However, AIM-FAS uses MAML~\cite{finn2017model} strategy for meta-learning and is orthogonal to our method. And thus the results cannot be compared fairly, so no comparisons are given here.

Besides, we also compare our results with two baselines. One is the source only ``$baseline$'', and the other is the joint training baseline ``$baseline_{joint}$'' by jointly training on the source and few target labeled samples, both of which are trained with cross-entropy loss. 
For fairly comparison, we re-implement CCSA and FADA based on our FAS framework because their original framework are not designed for FAS. 
When comparing with SSDG, we treat the original source domain and the few-shot target domain as two different source domains for domain generalization, and evaluate the performance of SSDG on the test sets of both domains. We maintain experiment settings of SSDG the same as the original paper.

\subsubsection{Comparison Results}

\begin{table}[t]
\centering

\caption{Experimental results on single-target domain benchmark \textbf{CMOS-ST}. (Notations: C: CelebA-Spoof; M: MSU-MFSD; O: Oulu-NPU; S: SiW)} 
\vspace{3pt}
\resizebox{0.45\textwidth}{!}{
\label{table:benchmark_1}
\begin{tabular}{cccccc}
\hline
                         &                          & \multicolumn{3}{c}{Source}                                  & Target                             \\ \cline{3-6} 
\multirow{-2}{*}{Protocols}  & \multirow{-2}{*}{Methods} & APCER(\%)↓ & BPCER(\%)↓ & ACER(\%)↓                         & HTER(\%)↓                          \\ \hline
                         & $baseline$              & 0.91       & 1.36       & 1.14                              & 25.68                              \\ \cline{2-6} 
                         & $baseline_{joint}$             & 0.96$\pm$0.11  & 1.47$\pm$0.08  & 1.22$\pm$0.08                         & 15.14$\pm$4.61                         \\
                         & CCSA                     & 3.57$\pm$0.71  & 1.24$\pm$0.47  & 2.4$\pm$0.31                          & 12.81$\pm$1.99                         \\
                         & FADA                     & 1.72$\pm$0.63  & 1.59$\pm$0.12  & 1.65$\pm$0.26                         & 14.07$\pm$2.07                         \\
                         & SSDG                     & 5.48$\pm$1.47  & 4.9$\pm$0.71   & 5.19$\pm$0.6                          & 20.68$\pm$3.15                         \\
\multirow{-6}{*}{C $\rightarrow$ M} & \textbf{ours}                     & 1.02$\pm$0.04  & 1.29$\pm$0.07  & \textbf{1.16}$\pm$\textbf{0.02} & \textbf{7.21}$\pm$\textbf{2.18}  \\ \hline
                         & $baseline$              & 0.91       & 1.36       & 1.14                              & 40.12                              \\ \cline{2-6} 
                         & $baseline_{joint}$             & 0.95$\pm$0.19  & 1.41$\pm$0.18  & 1.18$\pm$0.05                         & 25$\pm$2.65                            \\
                         & CCSA                     & 3.15$\pm$1.27  & 1.49$\pm$0.27  & 2.32$\pm$0.52                         & 20.87$\pm$1.81                         \\
                         & FADA                     & 1.83$\pm$0.41  & 1.71$\pm$0.29  & 1.77$\pm$0.1                          & 18.61$\pm$4.71                         \\
                         & SSDG                     & 6.11$\pm$1.12  & 5.26$\pm$1.01  & 5.68$\pm$0.91                         & 30.33$\pm$2.96                         \\
\multirow{-6}{*}{C $\rightarrow$ O} & \textbf{ours}                     & 0.9$\pm$0.07   & 1.41$\pm$0.07  & \textbf{1.15}$\pm$\textbf{0.03} & \textbf{11.85}$\pm$\textbf{0.89} \\ \hline
                         & $baseline$              & 0.91       & 1.36       & 1.14                              & 24.26                              \\ \cline{2-6} 
                         & $baseline_{joint}$             & 0.98$\pm$0.14  & 1.38$\pm$0.16  & \textbf{1.18}$\pm$\textbf{0.04}                         & 20.79$\pm$4.77                         \\
                         & CCSA                     & 3.7$\pm$1.27   & 1.26$\pm$0.49  & 2.48$\pm$0.43                         & 16.02$\pm$0.8                          \\
                         & FADA                     & 1.65$\pm$0.19  & 1.7$\pm$0.19   & 1.67$\pm$0.1                          & 18.22$\pm$4.33                         \\
                         & SSDG                     & 5.82$\pm$1.23  & 6.13$\pm$0.5   & 5.98$\pm$0.63                         & 17.84$\pm$1.28                         \\
\multirow{-6}{*}{C $\rightarrow$ S} & \textbf{ours}                     & 1.11$\pm$0.1   & 1.29$\pm$0.13  & 1.2$\pm$0.03  & \textbf{14.46}$\pm$\textbf{1.58} \\ \hline
\end{tabular} }
\vspace{-0.3em}
\end{table}
The experimental results on the proposed benchmark \textbf{CMOS-ST} are shown in Table~\ref{table:benchmark_1}. Compared with other SDA and DG-FAS methods, our method can significantly improve the performance on the target domain while maximally preserve the performance of the source domain using source data and only a few target training samples. Compared with $baseline$ and $baseline_{joint}$,  CCSA~\cite{motiian2017unified} and  FADA~\cite{motiian2017few} can 
make a certain improvement on the target domain, but sacrifice the performance on the source domain. The possible explanation is that the alignment methods they used may hurt the model's representative capability of the source domain data, so the model forgot the corresponding discriminative information. The results of SSDG~\cite{jia2020single} collapse in both the source domain and the target domain, because it is designed to learn shared knowledge from multiple labeled source domains with relatively balanced amounts of data and lacks a mechanism to deal with the serious imbalanced cases. 

\begin{table}[t]
\centering

\caption{Experimental results on multi-target domain benchmark \textbf{CMOS-MT}.}  
\vspace{3pt}
\resizebox{0.48\textwidth}{!}{
\label{table:benchmark_2}
\Huge
\begin{tabular}{cccccccc}
\hline
\multirow{2}{*}{Methods} & \multicolumn{3}{c}{Source}          & \multicolumn{4}{c}{Target}                                     \\ \cline{2-8} 
                        & APCER(\%)↓ & BPCER(\%)↓ & ACER(\%)↓ & HTER$_M$(\%)↓       & HTER$_O$(\%)↓        & HTER$_S$(\%)↓       & HTER$_{avg}$(\%)↓ \\ \hline
$baseline$       & 0.91       & 1.36       & 1.14      & 25.68              & 40.12               & 24.26         & 30.02      \\ \hline
$baseline_{joint}$      & 1.06±0.12  & 1.34±0.1   & 1.2±0.07  & 22.69±4.96         & 21.99±1.73          & 20.87±2.22     & 21.85±2.41     \\
CCSA                    & 3.72±0.34  & 1±0.14     & 2.36±0.18 & 12.06±1.51         & 22.75±2.77          & \textbf{13.25}±\textbf{1.23} & 16.02±1.55\\
FADA                    & 1.9±0.6    & 1.77±0.25  & 1.84±0.23 & 19.5±3.89          & 17.69±5.93          & 16.16±1.67     & 17.78±1.01     \\
SSDG                    & 7.15±1.98  & 5.87±1.03  & 6.51±1.35 & 19.07±3.49         & 22.97±4.25          & 16.34±6.68     & 19.46±4.25     \\
\textbf{ours}                    & 1.06±0.14  & 1.3±0.12   & \textbf{1.18}±\textbf{0.02} & \textbf{6.66}±\textbf{1.59} & \textbf{15.77}±\textbf{1.38} & 14.42±1.31     & \textbf{12.28}±\textbf{0.71}     \\ \hline

\end{tabular}
}
\vspace{-1em}
\end{table}

Results on multi-target benchmark \textbf{CMOS-MT} are shown in Table~\ref{table:benchmark_2}. Our method can consistently obtain considerable performance improvement when the available few target domain data is a mixture of multiple domains. Our results on SiW are slightly worse than CCSA, and the improvement on HTER is not as good as the other two protocols. Because the domain gap between SiW and CelebA-Spoof is actually larger than others, it is hard to achieve the optimal adaptation effect on the premise of maintaining the source performance.

\subsection{Ablation Studies}
To analyze the influence and interrelationship of each proposed module, we conduct ablation studies by using a fixed few-shot set in Oulu-NPU as the example target domain. The experimental results are shown in Table~\ref{table:ablation_add}
and~\ref{table:ablation_augment}.

\begin{table}[htb]
\centering
\caption{Quantitative results of the ablation studies. $X^{s}$ refers to source domain data, $X^{t}$ refers to target domain data, $X^{a}$ refers to augmented data. (a) Simply adding the $X^{a}$ into training can improve the performance on $X^{t}$. (b) $\mathcal{L}_{Lfc}$ can effectively suppress the negative impact of $X^{a}$ and $\mathcal{L}_{Cont}$ on the source domain. (c) $\mathcal{L}_{Cont}$ and $\mathcal{L}_{Adv}$ can make full use of the data from all of the three domains to improve the adaptability of the model on target individually and perform better when used simultaneously.} 
\vspace{3pt}
\resizebox{0.45\textwidth}{!}{
\label{table:ablation_add}
\begin{tabular}{lcccccccc}
\hline
                     & $X^{s}$ & $X^{t}$ & $X^{a}$ & $\mathcal{L}_{Lfc}$ & $\mathcal{L}_{Cont}$     & $\mathcal{L}_{Adv}$     & ACER(\%)↓ & HTER(\%)↓ \\ \hline
\multirow{3}{*}{(a)} & \checkmark  &    &    &          &                      &                      & 1.14      & 40.12     \\
                     & \checkmark  & \checkmark  &    &          &                      &                      & 1.20       & 22.63     \\
                     & \checkmark  & \checkmark  & \checkmark  &          &                      &                      & 1.37      & 17.38     \\ \hline
\multirow{4}{*}{(b)} & \checkmark  & \checkmark  & \checkmark  &          &                      &                      & 1.37      & 17.38     \\
                     & \checkmark  & \checkmark  & \checkmark  & \checkmark        & \multicolumn{1}{l}{} & \multicolumn{1}{l}{} & 1.15      & 18.21     \\
                     & \checkmark  & \checkmark  & \checkmark  &          & \checkmark                    & \multicolumn{1}{l}{} & 1.68      & 14.11     \\
                     & \checkmark  & \checkmark  & \checkmark  & \checkmark        & \checkmark                    & \multicolumn{1}{l}{} & 1.17      & 12.04     \\ \hline
\multirow{4}{*}{(c)} & \checkmark  & \checkmark  & \checkmark  & \checkmark        & \multicolumn{1}{l}{} & \multicolumn{1}{l}{} & 1.15      & 18.21     \\
                     & \checkmark  & \checkmark  & \checkmark  & \checkmark        & \checkmark                    & \multicolumn{1}{l}{} & 1.17      & 12.04     \\
                     & \checkmark  & \checkmark  & \checkmark  & \checkmark        &                      & \checkmark                    & 1.16      & 15.84     \\
                     & \checkmark  & \checkmark  & \checkmark  & \checkmark        & \checkmark                    & \checkmark                    & 1.17      & 10.32     \\ \hline
\end{tabular} }
\end{table}

\begin{table}[t]
\centering
\caption{Comparison with a general data augmentation method.} 
\vspace{3pt}
\resizebox{0.45\textwidth}{!}{
\label{table:ablation_augment}
\begin{tabular}{ccccc}
\hline
Aug methods          & APCER(\%)↓           & BPCER(\%)↓           & ACER(\%)↓            & HTER(\%)↓            \\ \hline
Random Erase~\cite{zhong2020random}         & 1.49                 & 1.12                 & 1.31                 & 16.48                \\
Style-based (ours)            & 0.86                 & 1.48                 & 1.17                 & 10.32                \\ \hline
\end{tabular} }
\vspace{-1em}
\end{table}

\noindent \textbf{Auxiliary Augmented Samples Benefit Target.}
As shown in Table~\ref{table:ablation_add} (a), when only sufficient source data are used for training, it is easy to achieve good results on source. However, the adaptability to the target domain is limited. As long as the training is sufficient, only few target data can actually improve the performance on $X^{t}$. However, the potential of the few labeled data are still not fully exploited. By simply adding the augmented data into joint training without mining the relationship of distributions, the performance on $X^{t}$ improves to a certain degree with slightly performance  drop on $X^{s}$. Because the augmented data enrich the training set of $X^{t}$,  but may introduce some noise to $X^{s}$. However, we argue that the proposed augmentation method already alleviates the issue of forgetting of source compared to existing augmentation methods. To verify our argument, we conduct the experiments by replacing our style transfer-based data augmentation method with a traditional one. The results are shown in Table~\ref{table:ablation_augment}, where an obvious decrease of performance can be observed.

\noindent \textbf{Source Domain Less-forgetting Constraint Ensures the Stability.}
Though the proposed augmentation method partially deals with the catastrophic forgetting on the source domain, the introduction of the augmented samples as well as domain alignment losses may still bring certain detrimental effects to the original feature representation of $X^{s}$, which are already discriminative and representative for the source domain. From Table~\ref{table:ablation_add} (b), it can be observed that the source domain less-forgetting constraint can further greatly alleviate the issue without losing much adaptability for $X^{t}$. Encouraging source features to stay in place can make the discriminative features of target move closer to source, instead of mapping to a new feature space that might be insufficient for either the source or the target.

\noindent \textbf{Complementary Effects of Instance-level and Distribution-level Alignments.}
Contrastive semantic alignment and progressive adversarial learning are two adopted strategies to improve adaptability on $X^{t}$, which work at instance-level and distribution-level, respectively. As shown in Table~\ref{table:ablation_add} (c), both methods can 
boost the target domain performance individually,
and the best results can be achieved when the two methods work together.

In addition to above analyses, we also conducted experiments on evaluating effectiveness of the progressive adversarial learning strategy. Compared to the results of non-progressive counterpart (11.94\%), around 2\% target domain performance gain is observed in the proposed strategy (10.32\%). With the help of our full method, we have brought more than 100\% performance gain on $X^{t}$ at the expense of only 2\% fluctuations on $X^{s}$.

\subsection{Visualizations and Analysis}
\begin{figure}[t]
\centering
\includegraphics[width=0.47\textwidth]{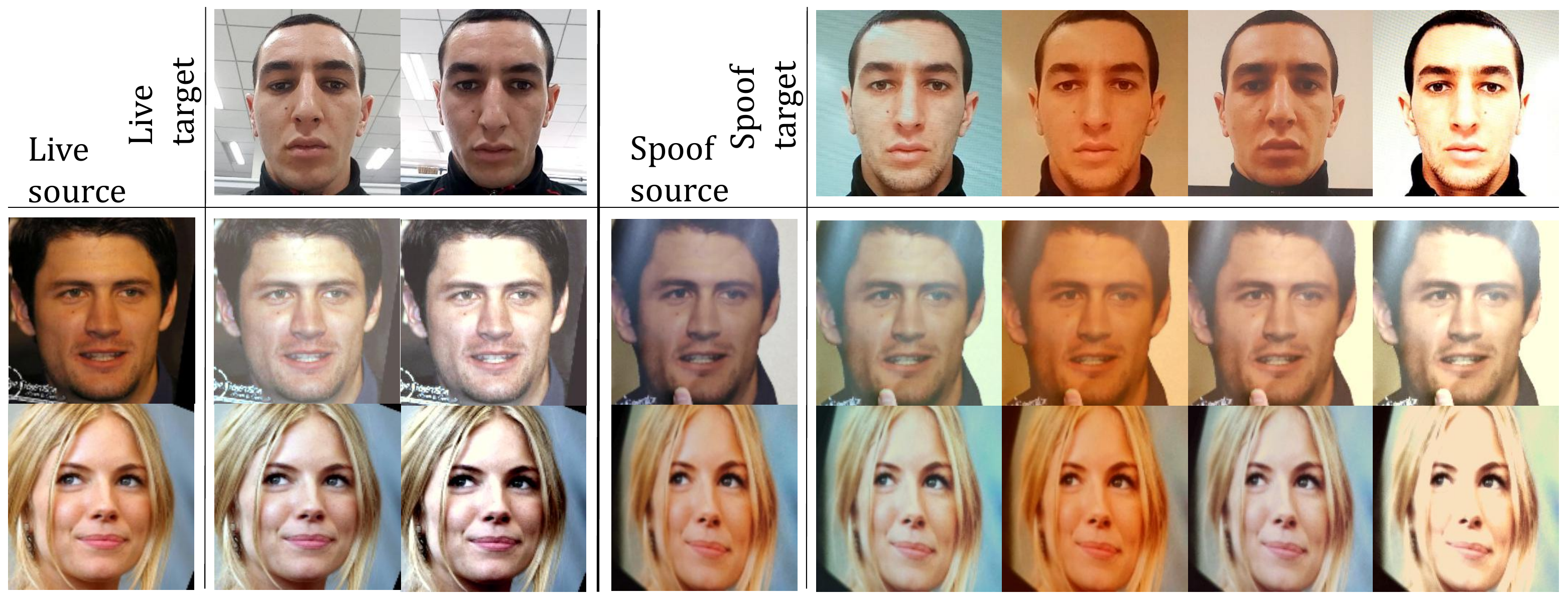}
\caption{
     Illustration of sample images of the augmented auxiliary data generated with WCT2~\cite{yoo2019photorealistic}.
}
\label{figure:wct2_more_samples}
\vspace{-1em}
\end{figure}
\noindent \textbf{Qualitative results.} Some sample augmented images generated via style transfer are shown in Figure~\ref{figure:wct2_more_samples}. Please refer to Supplementary for more example images. It can be observed that, the synthesized images preserve the source domain face identity and global arrangement. Moreover, the source domain spoof traces are also embodied in the hack images. By contrast, the global styles of the generated images are more consistent with the target domain. Thus, not only the global domain knowledge (lighting conditions, color/contrast related to camera parameters) is encoded, the spoof traces specific to the target domain are also reflected. Therefore, the augmented domain effectively consider the key knowledge in both the source and target domains, and smoothly bridge the two domains in the meantime. 

\begin{figure}[t]
\centering
\includegraphics[width=0.35\textwidth]{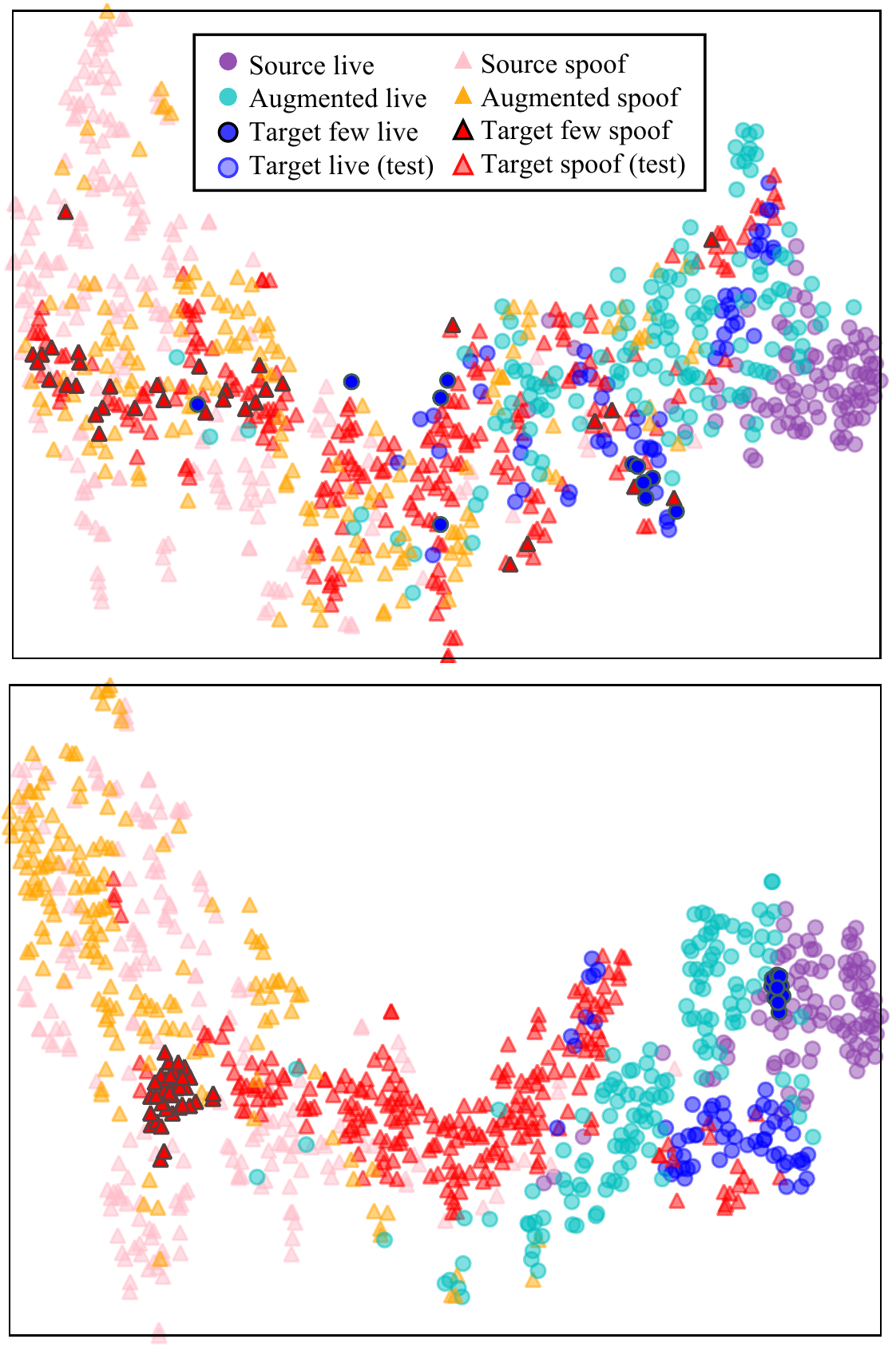}
\caption{
    The t-SNE~\cite{maaten2008visualizing} visualizations of the extracted features. Upper is pretrained by only source data. Lower is learned with our method, where the features are discriminative w.r.t all the domains.
}
\label{figure:tsne}
\vspace{-1em}
\end{figure}

\noindent \textbf{Distribution visualizations.} As shown in Figure~\ref{figure:tsne}, we select 128 samples from $X^{s}$, test set of $X^{t}$ and $X^{a}$ respectively as well as all the few labeled target samples for feature visualization using t-SNE~\cite{maaten2008visualizing}. As shown in the upper figure, the generated live and spoof style fusion data locate between source and target domain samples in the feature space, which can bridge domain gap smoothly. Also, the generated samples can somewhat simulate the distribution of unseen target test samples.
Without our method, the data of target domain are scattered in the whole feature space without clear decision boundary, which indicate that the feature generator provides week discriminant ability on target domain. As shown in the lower figure, contrastive semantic alignment pulls the cross domain samples of the same class closer while progressive adversarial learning strengthens distribution mixing. Thus, most of the target samples can be correctly classified. More t-SNE visualizations can be referred to Supplementary.

\section{Conclusion}
In this paper, we consider face anti-spoofing(FAS) in a more practical application scenario, i.e., when a few labeled new domain data can be obtained, how to improve the performance on the new target domain while ensuring the performance stability on existing source domain. 
To tackle this problem, we propose a Style transfer-based Augmentation for Semantic Alignment (SASA) framework, which create an auxiliary domain using photorealistic style transfer and then smoothly align the target data to the source domain by contrastive semantic alignment and progressive adversarial learning with the assistent of the generated auxiliary domain. Source domain less-forgetting constraint is introduced to further stabilize the performance on the source domain by penalizing drastic shift of the source samples in the feature space. To validate the performance of the proposed method, two benchmarks are proposed and the experimental results verified the effectiveness of SASA on few-shot domain expansion problem in face anti-spoofing tasks.

\clearpage

{\small
\bibliographystyle{ieee_fullname}
\bibliography{bibliography}
}

\end{document}